\documentclass{mva_style}
\usepackage{graphicx}
\usepackage{hyperref}
\usepackage{booktabs}
\usepackage{balance}

\finalcopy 

\begin{document}
\title{Small Object Detection for Birds with Swin Transformer}

\author{\\
  Da Huo\textsuperscript{1}, 
  Marc A. Kastner\textsuperscript{2}, Tingwei Liu\textsuperscript{1}, Yasutomo Kawanishi\textsuperscript{3,1}, Takatsugu Hirayama\textsuperscript{4,1},\\
  Takahiro Komamizu\textsuperscript{1}, Ichiro Ide\textsuperscript{1}
  \and
  \textsuperscript{1}Nagoya University\\ 
  Chikusa-ku, Nagoya, 464-8601, Japan\\
  \tt \{huod,liut\}@cs.is.i.nagoya-u.ac.jp,\\
  \tt {taka-coma}@acm.org,
  \tt {ide}@i.nagoya-u.ac.jp
  \and
  \textsuperscript{2}Kyoto University\\
  Sakyo-ku, Kyoto, 606-8501, Japan\\
  {\tt mkastner@i.kyoto-u.ac.jp}
  \and
  \textsuperscript{3}GRP, RIKEN\\ 
  Seika-cho, Kyoto 619-0288, Japan\\
  {\tt yasutomo.kawanishi@riken.jp}
  \and
  \textsuperscript{4}University of Human Environments\\ 
  Okazaki, Aichi, 444-3505, Japan\\
  {\tt t-hirayama@uhe.ac.jp}
}

\maketitle

\section*{\centering Abstract}
\textit{
Object detection is the task of detecting objects in an image. In this task, the detection of small objects is particularly difficult. Other than the small size, it is also accompanied by difficulties due to blur, occlusion, and so on. Current small object detection methods are tailored to small and dense situations, such as pedestrians in a crowd or far objects in remote sensing scenarios. However, when the target object is small and sparse, there is a lack of objects available for training, making it more difficult to learn effective features. In this paper, we propose a specialized method for detecting a specific category of small objects; birds. Particularly, we improve the features learned by the neck; the sub-network between the backbone and the prediction head, to learn more effective features with a hierarchical design. We employ Swin Transformer to upsample the image features. Moreover, we change the shifted window size for adapting to small objects. Experiments show that the proposed Swin Transformer-based neck combined with CenterNet can lead to good performance by changing the window sizes. We further find that smaller window sizes (default 2) benefit mAPs for small object detection.
}

\begin{figure}[!t]
  \begin{center}
  \includegraphics[width=0.4\textwidth,trim=50 600 280 50,clip]{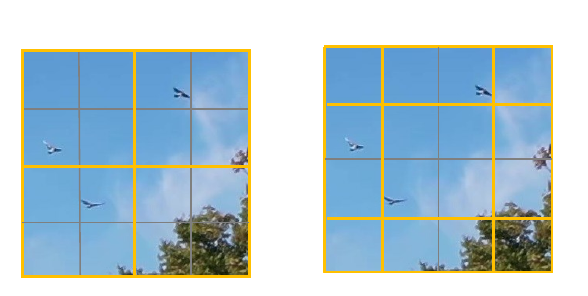}
  \end{center}
  \vspace{-1.5em}
  \qquad\qquad(a) layer \textit{l} \qquad\quad\quad\quad (b) layer \textit{l} + 1
   \vspace{0.5em}
  \caption{Illustration of the proposed shifted window self-attention approach in the Swin Transformer blocks in the neck network. In layer \textit{l}, a default window partitioning scheme is adopted (window size 2), and self-attention is computed inside each window. In the next layer \textit{l} + 1, the window partitioning is shifted by half of the window size (here, 1) for producing new windows. The birds in the windows of layer \textit{l} + 1 cross the boundaries of the previous windows of layer \textit{l}, provided cross window attention shifting with smaller window size for small object detection.}
  \label{figure1}
\end{figure}

\begin{figure*}[!ht]
  \begin{center}
   \includegraphics[width=1.0\textwidth]{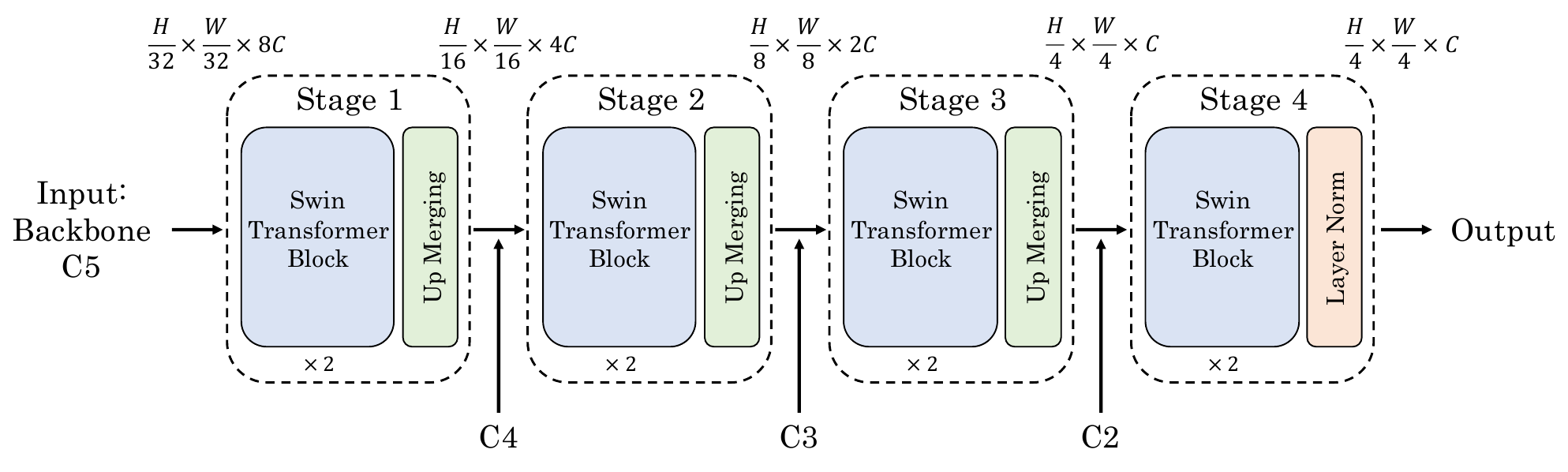}
   \end{center}
   \caption{Architecture of the proposed neck network. Feature C5 from the final stage of the backbone is the input. Three features with different scales, also from the backbone, namely C2, C3, and C4 are merged. The output is used for the final detection.}
   \label{figure2}
\end{figure*}

\section{Introduction}

With the development of deep learning, object detection has achieved good performance in various challenging benchmarks. It is successfully applied in many applications, such as autonomous driving, surveillance, and remote sensing. However, detecting small objects, particularly those smaller than 32 pixels, is still a challenging task compared to objects in general sizes~\cite{tinyimage2008, querydet2022}. For example, even recent object detection methods that show good performance on general objects are subject to a drop of nearly half on small objects in MS COCO~\cite{querydet2022, cocodet} detection. 

Such a large performance gap is mainly caused by the following factors: 1) Features of small objects disappear after many stages of down and up samplings, hence it is hard to learn the detection model, 2) The common receptive field on low-resolution feature maps may not match the size of small objects. 

In recent years, various methods have been proposed to solve limitations in the detection of particularly small objects. Super Resolution is used to recover the information of low-resolution objects, such as SOD-MTGAN~\cite{eccv2018sod}. They use a pre-trained detector to obtain object regions and then use a generator to generate corresponding super-resolution objects of an image. The discriminator is responsible for distinguishing whether the object is real or fake, as well as a detector to predict the
category and location of the object. These methods have difficulties in stability when training and are not efficient in detection tasks. Data-based methods are also considered to address small object detection. Here, the goal is to address the imbalance problems of a dataset, such as a pixel region of small objects not being large enough for training sufficient features. The use of data augmentation~\cite{cvpr2019} solves this problem by simply adding duplicate small objects. However, these methods only alleviate issues of specific datasets but do not address the difficulties in learning efficient features for small objects.

To sufficiently learn object features, we focus on the architecture of the object detection model for fitting the small object scale. In general, the common components of a detection model are: a backbone for image feature extraction, a neck with pyramid-like architecture combining different sizes of image feature maps for feature merging, and a prediction head for object classes and bounding box final prediction. 

In contrast to existing work using CNN-based networks, such as Feature Pyramid Networks (FPN)~\cite{lin2017feature}, we use Swin Transformer~\cite{swin2021} in both the backbone and the neck, which solves the previous limitation of the inconsistency of features, particularly while merging respective ones. Further, to adapt to the small object scale, we change the default window size in the neck to pay more attention to small objects, as shown in Figure~\ref{figure1}.

\section{Related Work}

\subsection{Object Detection Approaches}

{\noindent \bf {Anchor-based Approaches}}. A large number of methods use pre-defined anchors for object detection, such as YOLO v3~\cite{Yolov3}, which generates region proposals within the detection network. It samples pre-defined fixed-shape bounding boxes (called \textit{anchors}) and classifies each into “foreground or not”. An anchor is labeled foreground with $a$ {\textgreater} $0.7$ overlap with any ground-truth object as positive samples, background with $a$ {\textless} $0.3$ overlap as negative samples, or ignored otherwise. Each generated region proposal is again prepared for final prediction. However, in the case of small objects, the small scales leading to most of the anchors are selected as negative samples which causes an imbalance between positive ones while training the model. 

{\noindent \bf {Anchor-free Approaches}}. Currently, anchor-free methods are considered as one of the solutions for small object detection. CenterNet~\cite{zhou2019objects} only predicts the center points and the bounding boxes of the objects directly without IoU-based anchors for easily learning small objects. Further, they use focal loss to deal with the imbalance between positive and negative samples. Anchor-free methods are end-to-end, simple, fast, and also yield high performance in object detection. 
The proposed method builds on existing anchor-free methods, making them more robust through an improved neck network.

\subsection{Swin Transformer}

Recently, Swin Transformer~\cite{swin2021} has been proposed showing outstanding performance in most vision tasks, and quickly attracted significant attention. It shows great potential with hierarchical architecture for detecting objects with multiple scales. Further, they use shifting window-based attention that can make the transformer model learn features with local information by limiting the calculation of self-attention to a local window, greatly reducing the computational complexity. Swin Transformer can learn both local and global features, achieving excellent results. In the proposed method, we employ it as part of the backbone network and the neck network.


\subsection{Multi-scale Feature Learning}

The general idea is to learn objects in different scales by producing features with several sizes. The most famous method is the Feature Pyramid Network (FPN)~\cite{lin2017feature}. In deep neural networks, deep layers are generally rich in abstract semantic information, while shallow layers have more geometric details. The main idea is to integrate low-level spatial information and high-level semantic information to enhance object representation in multiple levels. Many methods~\cite{trident, effect} made improvements based on FPN for learning better representation in different scales for small object detection. The proposed method uses shifted windows for learning object features in several levels similar to FPN.

\section{Method}
\subsection{Overall Architecture}

The proposed method follows the common architecture with three parts: backbone, neck, and prediction head. In particular, we choose CenterNet~\cite{zhou2019objects} as a baseline for small object detection. In the backbone network, we use Swin Transformer~\cite{swin2021} for producing multiple levels of features instead of ResNet-50~\cite{resnet}.

In the neck, we propose a hierarchical network based on Swin Transformer rather than a CNN-based CenterNetNeck~\cite{zhou2019objects} or FPN~\cite{lin2017feature}. An overview of the proposed network is presented in Figure~\ref{figure2}, which illustrates a hierarchical architecture in each stage based on Swin Transformer. It first uses the final output of the Swin Transformer backbone into the first stage of the neck network. In our implementation, we adopt multiple (two as default) Swin Transformer blocks in each stage of the neck. In each stage, we also follow the Swin Transformer which calculates the attention inside of the window. This allows us to focus on more local information similar to convolution operation. Further, with the shifting windows in each stage, attentions of overlapping windows can be considered by the delicate shifting window design. Finally, we use the Up Merging module to upsample the feature maps in each stage to make features with multiple scales for better detection. 

\subsection{Shifted Windows Used in Neck Network}

The default window size of the Swin Transformer model is 7, contributing to excellent performance in general computer vision tasks, including object detection and semantic segmentation. It can detect objects in general sizes with the default window size. However, it may not be optimal to handle objects in small sizes. 

Here, we consider small objects as those less than 32 pixels. In such cases, after $32\times$ down-sampling, it can only represent one point of the feature map at most. If the window size were set to the default 7, the attention will be neglected inside such a big window. Moreover, with the shifted window approach for obtaining attention between windows, small object will not appear on different windows in the general object size. Instead, it will always be inside the same window; The cross-windows attention will not be considered in this situation. Here, the window partitioning scheme is shown in Figure~\ref{figure1}, visualizing the default window size of $2\times2$. By changing the window size to something smaller, the model considers the attention of the surrounding of small objects, such as the sky in the lower-left  of the bird of the \textit{l}-th layer, and the upper-right of the cloud around the bird in the \textit{l}+1-th layer. 

\subsection{Upmerging Blocks for Feature Upsampling}

\begin{figure}[t]
  \begin{center}
 \includegraphics[width=0.3\textwidth,trim=100 630 300 70,clip]{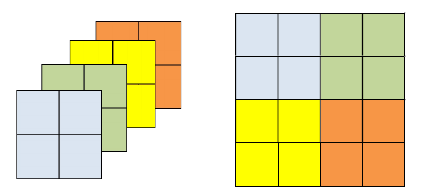}
  \end{center}
  \caption{Illustration of the Up Merging module of Swin Transformer neck for upsampling features. We use it in three stages for upsampling $2\times$ feature size.}
  \label{figure3}
\end{figure}

Between multiple stages of the neck network, it also must satisfy different sizes of feature maps for multiple scale detection. Considering the efficiency of the neck, we propose an upsampling module using simple operations instead of transposed convolution. The upsampling mechanism is the reverse of the Patch Merging for down-sampling; We call it Up Merging module, as shown in Figure~\ref{figure3}, which is also adopted in an image super-resolution method called PixelShuffle~\cite{upsample}. We set the stride of 2 for this process. After this, the number of channels of the feature maps becomes $C$ from the input channel (denoted as $4C$). To be consistent with the backbone for feature merging, we use a linear layer to change the channel from $C$ to $2C$.

\subsection{Skip Connections for Feature Merging}

With the long distance of both the backbone and neck networks, a large amount of information is lost through many down-sampling and up-sampling processes. Different from the CenterNet~\cite{zhou2019objects}, we take a U-Net~\cite{u-net}-like architecture and exploit skip-connections for feature merging to deal with long path losses and provide efficient location and object details from the backbone network. Here, we use concatenation with the features of the backbone and neck networks. 

\section{Experiments}
\subsection{Small Object Dataset}

We conduct experiments on two datasets for bird object detection, Drone2021~\cite{IMR}, which consists of 47,260 images with 60,971 annotated bird instances, and MVA2023~\cite{sodbchallenge2023misc}, which consists of 9,759 images with 29,037 annotated bird instances. Our experiments follow the baseline with CenterNet proposed by MVA2023~\cite{baselinecode_mva2023_sod_challenge}.

In the following, we first change the backbone network from ResNet50~\cite{resnet} to the current popularly used Swin Transformer (Swin-S)~\cite{swin2021} pre-trained with ImageNet22K~\cite{russakovsky2015imagenet}. We keep the CenterNet head with no change with three sub-networks to predict the center points, width, and height of the bounding box, and an offset for each center point, respectively. 

\subsection{Settings and Results}
We use training steps and settings as follows: first, we train the model on Drone2021 for 150 epochs with a learning rate of 5e$^{-6}$. Next, we fine-tune it on MVA2023 for 100 epochs with a learning rate of 1e$^{-5}$.
Thirdly, we generate predictions on MVA2023 to select hard-negatives examples. Next, we use hard negative training on MVA2023 with a hard-negative rate of 0.3 to introduce images for training. Here we change the setting in the head of CenterNet for better performance to make the gamma 6.0 in loss function for center point prediction and L1 loss weight 0.2 for the prediction of the bounding box. 
Finally, we generate the predictions for testing in the MVA challenge. As a result, we obtained AP{$_{50}$} of $0.702$ in the online public test that contains 9,699 images on the MVA challenge server.

\subsection{Ablation Study}
{\noindent \bf {Swin Transformer Neck:}} To explore the ability of the proposed network for detecting small objects, we compare the proposed Swin Transformer neck with the previous CenterNet neck on the AP performance. The results validated on MVA2023 validation set are shown in Table~\ref{table1}. We can see that the proposed Swin Transformer neck performs better for small-sized objects than CenterNetNeck used as default in CenterNet for general sized object detection.\\


\begin{table}[t]
  \caption{Mean Average Precision between CenterNet neck and Swin Transformer neck in MVA2023~\cite{sodbchallenge2023misc} validation set.}
  \begin{center}
    \begin{tabular}{c|ccc}
      \toprule
      Neck & AP$_{50}$ & AP$_{75}$ & AP\\
      \midrule
      CenterNet & 0.846   & 0.337 & 0.702 \\
      Swin Transformer & \bf{0.898} & \bf{0.372}& \bf{0.745}\\
      \bottomrule
    \end{tabular}
    \label{table1}
  \end{center}
\end{table}

\begin{figure}[!ht]
  \begin{center}
  \includegraphics[width=0.37\textwidth]{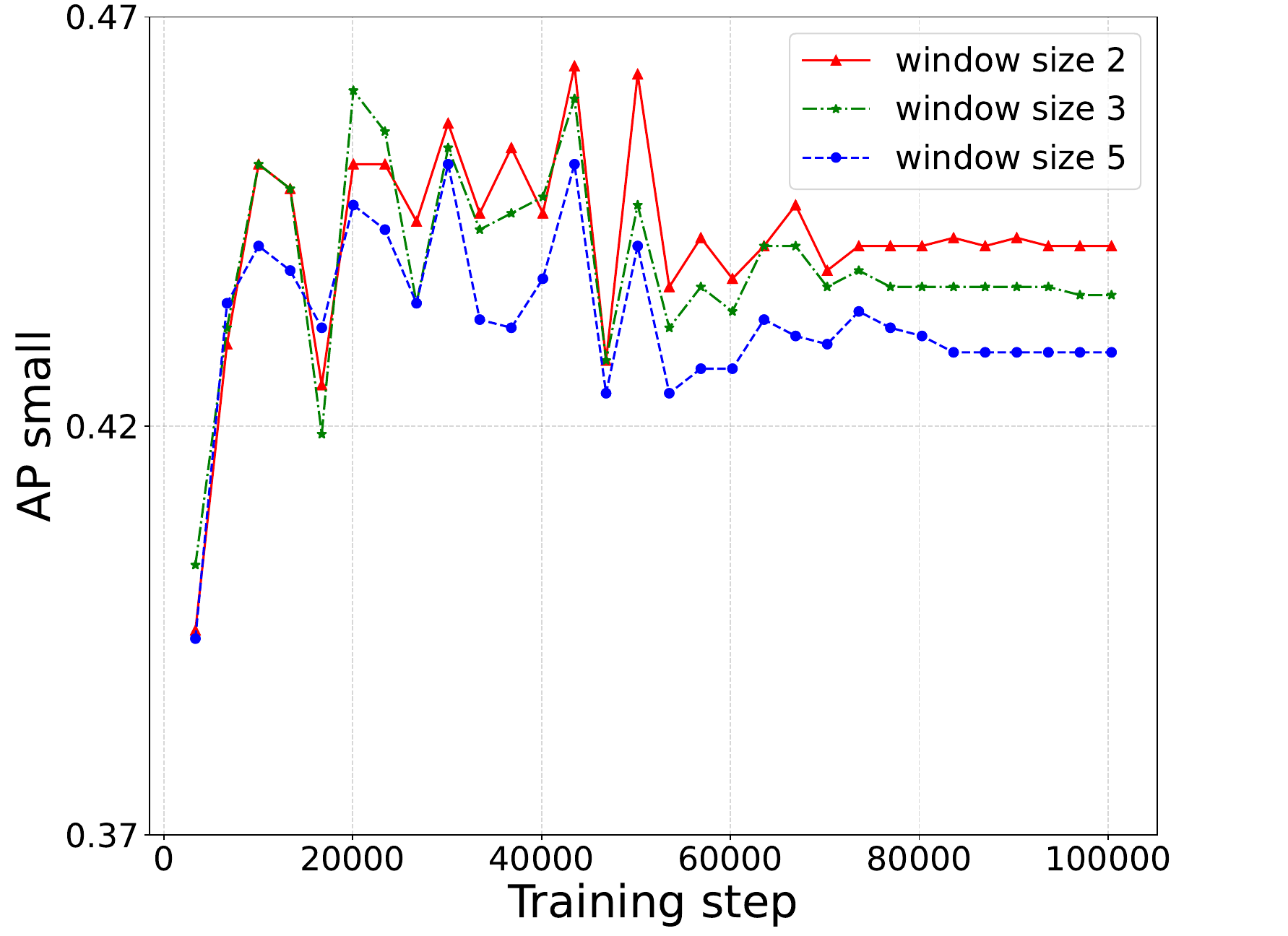}
  \end{center}
  \caption{Mean Average Precision of small objects under 32 pixels (AP{$_S$}) with different windows sizes 2, 3, and 5 of Swin Transformer neck, from Drone2021~\cite{IMR} validation set.}
  \label{sample-figure}
\end{figure}

{\noindent \bf {Changing the Window Size:}} Moreover, we explore the impact of the window size in the neck. Considering the training time, we report the performance before hard-negative training. The results shown in Table~\ref{table2} indicate that the smaller the window size, the better performance is obtained in all metrics.
Further, for small object under 32 pixels, the performance shows clearly that a smaller window size, especially window size 2, yields better AP{$_{S}$}, as shown in Figure~\ref{sample-figure}.


\begin{table}[t]
  \caption{Mean Average Precision in different window size of Swin Transformer neck in MVA2023~\cite{sodbchallenge2023misc} validation set, reported before the hard negative training.}
  \begin{center}
    \begin{tabular}{c|ccc}
      \toprule
      Window Size & AP$_{50}$ & AP$_{75}$ & AP\\
      \midrule
      2 & \bf{0.702} & \bf{0.171} & \bf{0.549} \\
      3 & 0.693   & 0.166 & 0.538 \\
      5 & 0.684 &  0.158  & 0.53  \\
      \bottomrule
    \end{tabular}
    \label{table2}
  \end{center}
\end{table}

\section{Conclusion}

We proposed a specialized method for detecting small objects.
Our contributions are as follows:
1) We proposed an hierarchical neck network based on Swin Transformer rather than CNN. 2) To adapt to small objects, we changed the default window size to a small one in the neck network to pay more attention to small objects. 3) Experimental results showed the effecticeness of the above two contributions in AP.

\section*{Acknowledgments}
Parts of the work presented in this paper were supported by the CSC / MEXT scholarship and MEXT Grant-in-aid for Scientific Research (20H00475), and the experiments used the “mdx: a platform for the data-driven future”~\cite{mdx-IEEECBDCom2022}.


\begin{thebibliography}{99}

\bibitem{tinyimage2008} A. Torralba, R. Fergus, and W. T. Freeman, “80 million tiny images: A large data set for nonparametric object and scene recognition,” \textit{IEEE Transactions on Pattern Analysis and
Machine Intelligence}, vol. 30, no. 11, pp.1958--1970, 2008.

\bibitem{querydet2022} C. Yang, Z. Huang, and N. Wang. ``QueryDet: Cascaded sparse query for accelerating high-resolution small object detection", \textit{Proc. 2022 IEEE/CVF Conference on Computer Vision and Pattern Recognition}, pp.13668--13677, 2022.

\bibitem{cocodet} T.-Y. Lin, M. Maire, S. Belongie, J. Hays, P. Perona, D. Ramanan, P. Doll{\'a}r, and C. L. Zitnick. 
``Microsoft COCO: Common objects in context", \textit{Proc. 13th European Conference on Computer Vision}, vol. 5, pp.740–755, 2014.

\bibitem{eccv2018sod} Y. Bai, Y. Zhang, M. Ding, and B. Ghanem,
``SOD-MTGAN: Small Object Detection via Multi-Task Generative Adversarial Network", \textit{Proc. 15th European Conference on Computer Vision}, vol. 13, pp.206–221, 2018.

\bibitem{cvpr2019} M. Kisantal, Z. Wojna, J. Murawski, J. Naruniec, and K. Cho,
``Augmentation for small object detection", \textit{Computing Research Reposiory arXiv Preprints, 2019}, arXiv:1902.07296.

\bibitem{lin2017feature} T. Lin, P. Dollár, R. Girshick, K. He, B. Hariharan, and S. Belongie,
``Feature pyramid networks for object detection", \textit{Proc. 2017 IEEE Conference on Computer Vision and Pattern Recognition}, pp.936--944, 2017.

\bibitem{Yolov3} J. Redmon and A. Farhadi, 
``YOLOv3: An incremental improvement", \textit{Computing Research Reposiory arXiv Preprints, 2018}, arXiv:1804.02767.

\bibitem{zhou2019objects} X. Zhou, D. Wang, and P. Krähenbühl, 
``Objects as points", \textit{Computing Research Reposiory arXiv Preprints, 2019}, arXiv:1904.07850, .

\bibitem{swin2021} Z. Liu, Y. Lin, Y. Cao, H. Hu, Y. Wei, Z. Zhang, S. Lin, and B. Guo, 
``Swin Transformer: Hierarchical vision transformer using shifted windows", \textit{Proc. 18th IEEE International Conference on Computer Vision}, pp.9992--10002, 2021.

\bibitem{trident} Y. Li, Y. Chen, N. Wang, and Z. Zhang, 
``Scale-aware trident networks for object detection", \textit{Proc. 17th IEEE International Conference on Computer Vision}, pp.6053--6062, 2019.

\bibitem{effect}Y. Gong, X. Yu, Y. Ding, X. Peng, J. Zhao, and Z. Han,
``Effective fusion factor in FPN for tiny object detection", \textit{Proc. 21st Winter Conference on Applications of Computer Vision}, pp.1159--1167, 2021.

\bibitem{resnet} B. Xiao, H. Wu, and Y. Wei, 
``Simple baselines for human pose estimation and tracking", \textit{Proc. 15th European Conference on Computer Vision}, vol. 6, pp.472--487, 2018.

\bibitem{upsample} W. Shi, J. Caballero, F. Huszár, J. Totz, , A. P. Aitken, R. Bishop, D. Rueckert, and Z. Wang, ``Real-time single image and video super-resolution using an efficient sub-pixel convolutional neural network", \textit{Proc. 2016 IEEE Conference on Computer Vision and Pattern Recognition}, pp.1874--1883, 2016.


\bibitem{u-net}O. Ronneberger, P. Fischer, and T. Brox,
``U-Net: Convolutional networks for biomedical image segmentation", \textit{Proc. 18th Medical Image Computing and Computer Assisted Intervention Conference}, vol. 3, pp.234--241, 2015.




\bibitem{IMR}
  S. Fujii, K. Akita, and N. Ukita: 
  ``Distant bird detection for safe drone flight and its dataset'', 
  \textit{Proc. 17th International Conference on Machine Vision and Applications}, pp.1--5, 2021.

\bibitem{sodbchallenge2023misc}
{Y. Kondo, N. Ukita, and T. Yamaguchi},
{``MVA2023 Small object detection challenge for spotting birds"}, 
{\tt{Accessed: https://www.mva-org.jp/mva2023/\\challenge}}, {2023.}

\bibitem{baselinecode_mva2023_sod_challenge} K. Zhao, R. Miyata, Y. Kondo, and K. Akita, ``Baseline code for SOD4SB by IIM-TTIJ",\\{\tt{Accessed: https://github.com/IIM-TTIJ/MVA2023\\
SmallObjectDetection4SpottingBirds}}, {2023.}

\bibitem{russakovsky2015imagenet}
  O. Russakovsky, J. Deng,  H. Su, J. Krause, S. Satheesh, S. Ma, Z. Huang, A. Karpathy, A. Khosla, and M. Bernstein: 
  ``ImageNet large scale visual recognition challenge'', 
  \textit{International Journal of Computer Vision}, vol. 115, pp.211--252, 2015.

\bibitem{mdx-IEEECBDCom2022}
  T. Suzumura, A. Sugiki, H. Takizawa, A. Imakura, H. Nakamura, K. Taura, T. Kudoh, T. Hanawa, Y. Sekiya, H. Kobayashi, Y. Kuga, R. Nakamura, R. Jiang, J. Kawase, M. Hanai, H. Miyazaki, T. Ishizaki, D. Shimotoku, D. Miyamoto, K. Aida, A. Takefusa, T. Kurimoto, K. Sasayama, N. Kitagawa, I. Fujiwara, Y. Tanimura, T. Aoki, T. Endo, S. Ohshima, K. Fukazawa, S. Date, and T. Uchibayashi, ``mdx: A Cloud Platform for Supporting Data Science and Cross-Disciplinary Research Collaborations", \textit{Computing Research Reposiory arXiv Preprints, 2022}, arXiv:2203.14188.







  

\end{thebibliography}
\end{document}